\documentclass[10pt,twocolumn,letterpaper]{article}
\usepackage{pifont}
\usepackage{cvpr}
\usepackage{times}
\usepackage{epsfig}
\usepackage{graphicx}
\usepackage{amsmath}
\usepackage{amssymb}
\usepackage[ruled,vlined]{algorithm2e}
\usepackage{booktabs}
\usepackage{color}
\usepackage{multirow}
\usepackage{amsmath}
\usepackage{bm}
\usepackage{url}
\usepackage{bbding}
\usepackage{indentfirst}
\usepackage{bbm}
\usepackage[T1]{fontenc}
\usepackage{cite}
\usepackage{epstopdf}
\usepackage{hyperref}

\cvprfinalcopy 

\def\cvprPaperID{260} 

\ifcvprfinal\fi
\begin{document}

\title{COIN: A Large-scale Dataset for Comprehensive Instructional Video Analysis}

\author{Yansong Tang$^{1}$ \qquad 
Dajun Ding$^{2}$ \qquad 
Yongming Rao$^{1}$ \qquad 
Yu Zheng$^{1}$ \qquad 
\\
Danyang Zhang$^{1}$ \qquad 
Lili Zhao$^{2}$ \qquad 
Jiwen Lu$^{1}$\thanks{the corresponding author is Jiwen Lu.} \qquad 
Jie Zhou$^{1}$\\
\noindent
$^{1}$Department of Automation, Tsinghua University \qquad  
$^{2}$ Meitu Inc.\\
\url{https://coin-dataset.github.io/}
}

\maketitle

\begin{abstract}
There are substantial instructional videos on the Internet, 
which enables us to acquire knowledge for completing various tasks.
However, most existing datasets for instructional video analysis have the limitations in diversity and scale,
which makes them far from many real-world applications where more diverse activities occur.
Moreover, it still remains a great challenge to organize and harness such data.
To address these problems,
we introduce a large-scale dataset called ``COIN'' for COmprehensive INstructional video analysis.
Organized with a hierarchical structure, the COIN dataset contains 11,827 videos of 180 tasks in 12 domains (e.g., vehicles, gadgets, etc.) related to our daily life.
With a new developed toolbox,
all the videos are annotated effectively with a series of step descriptions and the corresponding temporal boundaries.
Furthermore, we propose a simple yet effective method to capture the dependencies among different steps,
which can be easily plugged into conventional proposal-based action detection methods for localizing important steps in instructional videos.
In order to provide a benchmark for instructional video analysis,
we evaluate plenty of approaches on the COIN dataset under different evaluation criteria.
We expect the introduction of the COIN dataset will
promote the future in-depth research on instructional video analysis for the community.
\end{abstract}


\section{Introduction}
Instructional videos provide intuitive visual examples for learners to acquire knowledge to accomplish different tasks.
With the explosion of video data on the Internet, people around the world have
uploaded and watched substantial instructional videos~\cite{DBLP:conf/cvpr/AlayracBASLL16,Sener_2015_ICCV}, covering miscellaneous categories.
According to the scientists in educational psychology~\cite{Nadolski2005Optimizing},
novices often face difficulties in learning from the whole realistic task,
and it is necessary to divide the whole task into smaller segments or steps as a form of simplification.
Actually, a variety of video analysis tasks in computer vision (e.g., action temporal localization~\cite{DBLP:conf/iccv/ZhaoXWWTL17, DBLP:conf/iccv/XuDS17}, video summarization~\cite{DBLP:journals/pami/ElhamifarSS16,DBLP:conf/eccv/ZhangCSG16,DBLP:journals/tip/PandaMR17} and video caption~\cite{DBLP:journals/corr/abs-1804-00819,DBLP:conf/iccv/KrishnaHRFN17,Yu_2018_CVPR}, etc) have been developed recently. 
Also, growing efforts have been devoted to exploiting different challenges of instructional video analysis~\cite{DBLP:conf/cvpr/HuangLFN17,DBLP:conf/aaai/ZhouXC18,Sener_2015_ICCV,DBLP:conf/cvpr/AlayracBASLL16}.

\begin{figure*}[tb]
\includegraphics[width = \linewidth]{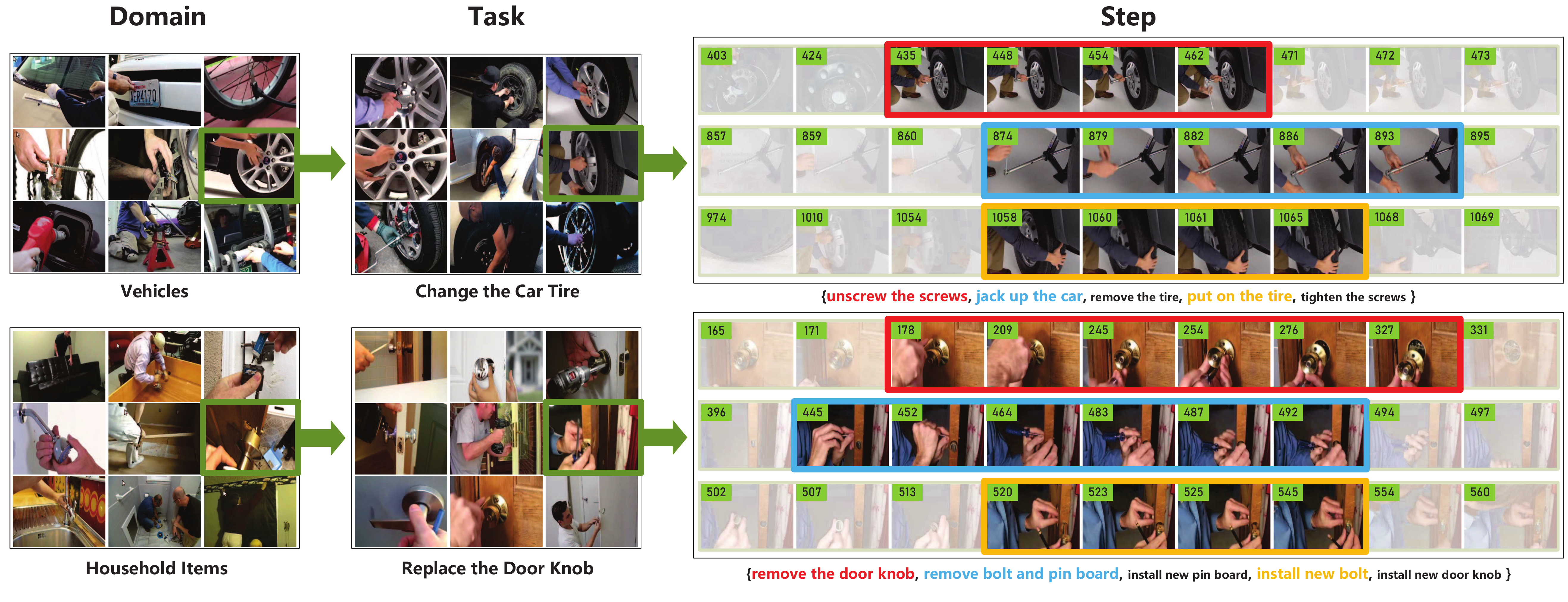}
\caption{Visualization of two root-to-leaf branches of the \textit{COIN}.
There are three levels of our dataset: \textit{domain}, \textit{task} and \textit{step}.
Take the top row as an example, 
in the left box, we show a set of frames of 9 different \textit{tasks} associated with the \textit{domain} ``vehicles''.
In the middle box, we present several images of 9 videos belonging to the \textit{task} ``change the car tire''.
Based on this \textit{task}, in the right box,
we display a sequence of frames sampled from a specific video,
where the indices are presented at the left-top of each frame.
The intervals in red, blue and yellow indicate the \textit{step} of ``unscrew the screws'', ``jack up the car'' and ``put on the tire'',
which are described with the text in corresponding color at the bottom of the right box.
All figures are best viewed in color.
}
\end{figure*}

\begin{table*}[tb]
\footnotesize
\begin{center}
\caption{\small Comparisons of existing instructional video datasets.}
\begin{tabular}{ r c c c c c c c c c }
\toprule[1.5pt]
 \textbf{Dataset} & \textbf{Duration} & \textbf{Samples}  & \textbf{Segments} &  \textbf{Type of Task} & \textbf{Video Source} & \textbf{Hierarchical}  & \textbf{Classes} & \textbf{Year}\\
 \midrule
  MPII\cite{DBLP:conf/cvpr/RohrbachAAS12} & 9h,48m & 44 & 5,609 & cooking activities & self-collected &  \ding{55}& - & 2012 \\ 
  YouCook\cite{DBLP:conf/cvpr/DasXDC13} & 2h,20m & 88 & - & cooking activities &  YouTube &\ding{55} &  - & 2013 \\
  50Salads\cite{DBLP:conf/huc/SteinM13} & 5h,20m & 50 & 966 & cooking activities &  self-collected &  \ding{55}   & - & 2013 \\
  Breakfast\cite{DBLP:conf/cvpr/KuehneAS14} & 77h & 1,989 & 8,456 & cooking activities &  self-collected & \ding{55} &  10 & 2014 \\
  "5 tasks"\cite{DBLP:conf/cvpr/AlayracBASLL16} & 5h & 150 & - & comprehensive tasks &  YouTube & \ding{55}  &  5 & 2016  \\
  Ikea-FA\cite{DBLP:conf/dicta/ToyerCHG17} & 3h,50m & 101 & 1,911 & assembling furniture & self-collected & \ding{55} & - & 2017  \\
  YouCook2\cite{DBLP:conf/aaai/ZhouXC18} & 176h & 2,000 & 13,829 & cooking activities & YouTube & \ding{55}   & 89 & 2018  \\
  EPIC-KITCHENS\cite{Damen_2018_ECCV}   & 55h & 432 & 39,596 & cooking activities & self-collected & \ding{55}   & - & 2018  \\  
\midrule
  \textbf{COIN} (Ours) & \textbf{476h,38m} & \textbf{11,827} & \textbf{46,354} & \textbf{comprehensive tasks} &  \textbf{YouTube} & \ding{51} & \textbf{180} & \\
\bottomrule[1.5pt]
\end{tabular}
\end{center}
\label{tabstata}
\end{table*}

In recent years, a number of datasets for instructional video analysis~\cite{DBLP:conf/cvpr/RohrbachAAS12, DBLP:conf/cvpr/DasXDC13, DBLP:conf/huc/SteinM13,DBLP:conf/cvpr/KuehneAS14,DBLP:conf/cvpr/AlayracBASLL16,DBLP:conf/dicta/ToyerCHG17,DBLP:conf/aaai/ZhouXC18}
have been collected in the community.
Annotated with texts and temporal boundaries of a series of steps to complete different tasks,
these datasets have provided good benchmarks for preliminary research.
However, most existing datasets focus on a specific domain like cooking, 
which makes them far from many real-world applications where more diverse activities occur.
Moreover, 
the scales of these datasets are insufficient
to satisfy the hunger of recent data-driven learning methods.

To tackle these problems, 
we introduce a new dataset called ``COIN'' for COmprehensive INstructional video analysis.
The COIN dataset contains 11,827 videos of 180 different tasks, 
covering the daily activities related to vehicles, gadgets, etc. 
Unlike existing instructional video datasets,
COIN is organized in a three-level semantic structure.
Take the top row of Figure 1 as an example,
the first level of this root-to-leaf branch is a \textit{domain} named ``vehicles",
under which there are numbers of video samples belonging to the second level \textit{tasks}.
For a specific task like ``change the car tire", 
It is comprised of a series of \textit{steps} such as ``unscrew the screws", ``jack up the car", ``put on the tire", etc.
These \textit{steps} appear in different interval of a video,
which belongs to the third-level tags of COIN.
We also provide the temporal boundaries of all the steps,
which are effectively annotated based on a new developed toolbox.

As another contribution,
we propose a new task-consistency method to localize different steps in instructional videos by considering their intrinsic dependencies.
First, as a bottom-up strategy,
we infer the task label of the whole video according to the prediction scores,
which can be obtained by existing proposal-based action detection methods. 
Then, as a top-down scheme,
we refine the proposal scores based on the predicted task label.
In order to set up a benchmark, 
we implement various approaches on the COIN. 
The experimental results have shown the great challenges of our dataset and the effectiveness of the proposed method for step localization.

\section{Related Work}
\textbf{Tasks for Instructional Video Analysis:}
There are various tasks for instructional video analysis, 
e.g., step localization, action segmentation, procedure segmentation~\cite{DBLP:conf/aaai/ZhouXC18}, dense video caption~\cite{DBLP:journals/corr/abs-1804-00819} and visual grounding~\cite{Huang_2018_CVPR,DBLP:conf/bmvc/ZhouLC18}.
In this paper, we focus on the first two tasks, where 
\textit{step localization} aims to  localize the start and end points of a series of steps and recognizing their labels, 
and \textit{action segmentation} targets to parse a video into different actions at frame-level.

\textbf{Datasets Related to Instructional Video Analysis:}
There are mainly three types of related datasets.
(1) The action detection datasets are comprised of untrimmed video samples,
and the goal is to recognize and localize the action instances on temporal domain~\cite{DBLP:conf/cvpr/HeilbronEGN15, THUMOS14} or spatial-temporal domain~\cite{Gu_2018_CVPR}.
(2) The video summarization datasets~\cite{DBLP:journals/prl/AvilaLLA11,DBLP:conf/eccv/GygliGRG14,DBLP:conf/cvpr/SongVSJ15,DBLP:journals/tip/PandaMR17} contain long videos arranging from different domains. The tasks is to extract a set of informative frames in order to briefly summarize the video content.
(3) The video caption datasets are annotated with descried sentences or phrases,
which can be based on either a trimmed video~\cite{DBLP:conf/cvpr/XuMYR16, Yu_2018_CVPR} or different segments of a long video~\cite{DBLP:conf/iccv/KrishnaHRFN17}.
Our COIN is relevant to the above mentioned datasets, 
as it requires to localize the temporal boundaries of important steps corresponding to a task.
The main differences lie in the following two aspects:
(1) \textbf{Consistency.} 
The steps belonging to different tasks shall not appear in the same video.
For example, it is unlikely for an instructional video to contain the step ``pour water to the tree'' (belongs to task ``plant tree'') and the step ``install the lampshade'' (belongs to task ``replace a bulb'').
(2) \textbf{Ordering.}
There may be some intrinsic ordering constraints among a series of steps for completing different tasks. 
For example, for the task of ``planting tree",
the step ``dig a hole" shall be ahead of the step ``put the tree into the hole".

There have been a variety of instructional video datasets proposed in recent years.
Table I summarizes the comparison among some publicly relevant instructional datasets and our proposed COIN.
While the existing datasets present various challenges to some extent, they still have some limitations in the following two aspects.
(1) \textbf{Diversity:} Most of these datasets tend to be specific and contain certain types of instructional activities, e.g., cooking.
However, according to some widely-used websites~\cite{howcast, howdini, wikihow}, people attempt to acquire knowledge from various types of instructional video across different domains.
(2) \textbf{Scale:}
Compared with the recent datasets for image classification (e.g., ImageNet~\cite{DBLP:conf/cvpr/DengDSLL009} with ~1 million images) and action detection (e.g., ActivityNet v1.3~\cite{DBLP:conf/cvpr/HeilbronEGN15} with ~20k videos),
most existing instructional video datasets are relatively smaller in scale.
The challenge to build such a large-scale dataset mainly stems from the difficulty to organize enormous amount of video and the heavy workload of annotation. 
To address these two issues, 
we first establish a rich semantic taxonomy covering 12 domains and collect 11,827 instructional videos to construct COIN. 
With our new developed toolbox,
we also provide the temporal boundaries of steps that appear in all the videos with effective and precise annotations.

\textbf{Methods for Instructional Video Analysis:}
The approaches for instructional video analysis can be roughly divided into three categories: unsupervised learning-based, weakly-supervised learning-based and fully-supervised learning-based.
For the first category,
the step localization task usually takes a video and the corresponding narration or subtitle as multi-modal inputs~\footnote{The language signal should not be treated as supervision since the steps are not directly given, but need to be further explored in an unsupervised manner.}.
For example,
Sener \textit{et al.}\cite{Sener_2015_ICCV} developed a joint generative model to parse both video frames and subtitles into activity steps.
Alayrac \textit{et al.}\cite{DBLP:conf/cvpr/AlayracBASLL16} leveraged the complementary nature of the instructional video and its narration
to discover and locate the main steps of a certain task.
Generally speaking, the advantages to employ the narration or subtitle is to avoid human annotation, which may cost huge workload.
However, these narration or subtitles may be inaccurate~\cite{DBLP:conf/aaai/ZhouXC18} or even irrelevant to the video\footnote{For example, in a video with YouTube ID CRRiYji\_K9Q,
the instructor talks a lot about other things when she performs the task ``injection''.
}.

For the second category,
Kuehne et al.~\cite{DBLP:conf/cvpr/KuehneAS14} developed a hierarchical model based on HMMs and a context-free grammar to parse the main steps in the cooking activities. Richard \textit{et al.}~\cite{richard2017action}\cite{DBLP:journals/corr/abs-1805-06875} adopted Viterbi algorithm to solve the probabilistic model of weakly supervised segmentation. Ding \textit{et al.}~\cite{DBLP:journals/corr/abs-1803-10699} proposed a temporal convolutional feature pyramid network to predict frame-wise labels and use soft boundary assignment to iteratively optimize the segmentation results.
In this work, we also evaluate these three methods\footnote{The details of the weak supervisions are described in section 5.2.} to provide a benchmark results on COIN.

For the third category, we focus on step localization.
This task is related to the area of action detection,
where promising progress has also been achieved recently.
For example,
Zhao \textit{et al.}~\cite{DBLP:conf/iccv/ZhaoXWWTL17} developed structured segment networks (SSN) to model the temporal structure of each action instance with a structured temporal pyramid.
Xu \textit{et al.}~\cite{DBLP:conf/iccv/XuDS17} introduced a Region Convolutional 3D Network (R-C3D) architecture,
which was built on C3D~\cite{DBLP:conf/iccv/TranBFTP15} and Faster R-CNN~\cite{faster-rcnn},
to explore the region information of video frames.
Compared with these methods,
we attempt to further explore the dependencies of different steps, which lies in the intrinsic structure of instructional videos.
Towards this goal, we proposed a new method with a bottom-up strategy and a top-down scheme.
Our method can be easily plugged into recent proposal-based action detection methods and enhance the performance of step localization in instructional videos.

\section{The COIN Dataset}
In this section we present COIN, 
a video-based dataset which covers an extensive range of everyday tasks with explicit steps. 
To our best knowledge,
it is currently the largest dataset for comprehensive instructional video analysis.
We will introduce COIN from the following aspects:
the establishment of lexicon, a new developed toolbox for efficient annotation, and the statistics of our dataset.

\begin{figure}[tb]
\includegraphics[width = \linewidth]{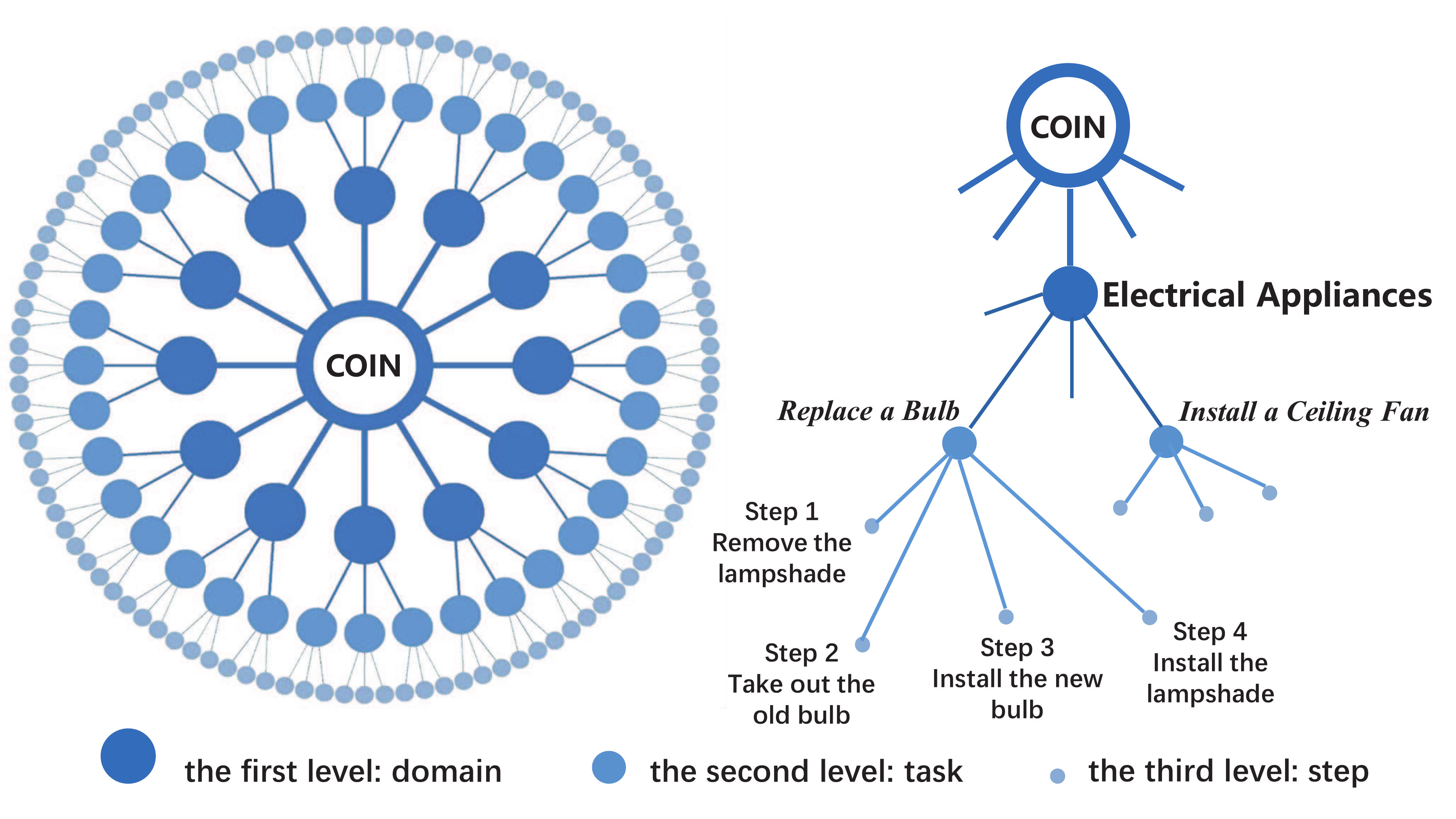}
\caption{Illustration of the COIN lexicon.
The left figure shows the hierarchical structure, 
where the nodes of three different sizes correspond to the domain, task and step respectively. For brevity, we do not draw all the tasks and steps here.
The right figure presents detailed steps of the task ``replace a bulb",
which belongs to the domain ``electrical appliances".
}
\end{figure}
%

\textbf{Lexicon:}
The purpose of COIN is to establish a rich semantic taxonomy to organize comprehensive instructional videos.
In previous literature,
some representative large-scale datasets were built upon existing structures.
For example, 
the ImageNet~\cite{DBLP:conf/cvpr/DengDSLL009} database was constructed based on a hierarchical structure of WordNet~\cite{WordNet},
while the ActivityNet dataset~\cite{DBLP:conf/cvpr/HeilbronEGN15} adopted the activity taxonomy organized by American Time Use Survey (ATUS)~\cite{ATUS}.
In comparison, 
it remains great difficulty to define such a semantic lexicon for instructional videos because of their high diversity and complex temporal structure.
Hence,
most existing instructional video datasets~\cite{DBLP:conf/aaai/ZhouXC18} focus on a specific domain like cooking or furniture assembling, 
and ~\cite{DBLP:conf/cvpr/AlayracBASLL16} only consists of five tasks.

Towards the goal of constructing a large-scale benchmark with high diversity, 
we proposed a hierarchical structure to organize our dataset.
Figure 1 and Figure 2 present the illustration of our lexicon,
which contains three levels from roots to leafs: domain, task and step.

(1) \textit{Domain.}
For the first level,
we bring the ideas from the organization of several websites\cite{howcast}\cite{wikihow}\cite{howdini},
which are commonly-used for users to watch or upload instructional videos.
We choose 12 domains as: \textit{nursing \& caring}, \textit{vehicles}, \textit{leisure \& performance}, \textit{gadgets}, \textit{electric appliances}, \textit{household items}, \textit{science \& craft}, \textit{plants \& fruits}, \textit{snacks \& drinks}, \textit{dishes}, \textit{sports}, and \textit{housework}.

(2) \textit{Task.}
As the second level, the task is linked to the domain.
For example, the tasks ``replace a bulb'' and ``install a ceiling fan'' are associated with the domain ``electrical appliances''.
As most tasks on \cite{howcast}\cite{wikihow}\cite{howdini} may be too specific,
we further search different tasks of the 12 domains on YouTube.
In order to ensure the tasks of COIN are commonly used,
we finally select 180 tasks, under which the searched videos are often viewed~\footnote{We present the statistics of browse times in supplementary material.}.

(3) \textit{Step.}
The third level of the lexicon are various series of steps to complete different tasks.
For example,
steps ``remove the lampshade'', ``take out the old bulb'', ``install the new bulb'' and ``install the lampshade'' are associated with the tasks ``replace a bulb''.
We employed 6 experts (e.g. driver, athlete, etc.) who have prior knowledge in the 12 domains to define these steps.
They were asked to browse the corresponding videos as a preparation in order to provide the high-quality definition,
and each step phrase will be double checked by another expert.
In total, there are 778 defined steps, 
where there are 4.84 words per phrase for each step.
Note that we do not directly adopt the narrated information,
which might have large variance for a specific task,
because we expect to obtain the simplification of the core steps,
which are common in different videos of accomplishing a certain task.
%

\textbf{Annotation Tool:}
Given an instructional video,
the goal of annotation is to label the step categories and the corresponding segments.
As the segments are variant in length and content, 
it will cost huge workload to label the COIN with conventional annotation tool.
In order to improve the annotation efficiency, we have developed a new toolbox which has two modes: \textit{frame mode} and \textit{video mode}.
Figure 4 shows an example interface of the \textit{frame mode}, which presents the frames extracted from a video under an adjustable frame rate (default is 2fps).
Under the \textit{frame mode}, the annotator can directly select the start and end frame of the segment as well as its label. 
However, 
due to the time gap between two adjacent frames,
some quick and consecutive actions might be missed.
To address this problem,
we adopted another \textit{video mode}.
The \textit{video mode} of the annotation tool presents the online video and timeline, which is frequently used in previous video annotation systems~\cite{DBLP:conf/iccv/KrishnaHRFN17}. 
Though the \textit{video mode} brings more continuous information in the time scale, 
it is much more time-consuming than the \textit{frame mode} because of the process to locate a certain frame and adjust the timeline\footnote{
For a set of videos, the annotation time under the \textit{frame mode} is only 26.8\% of that under the \textit{video mode}. Please see supplementary material for details.}.

During the annotation process,
each video is labelled by three different workers with payments.
To begin with, the first worker generated primary annotation under the \textit{frame mode}. 
Next, the second worker adjusted the annotation based on the results of the first worker. 
Ultimately, the third worker switched to the \textit{video mode} to check and refine the annotation.
Under this pipeline,
the total time of the annotation process is about 600 hours.

\begin{figure}[tb]
\includegraphics[width = \linewidth]{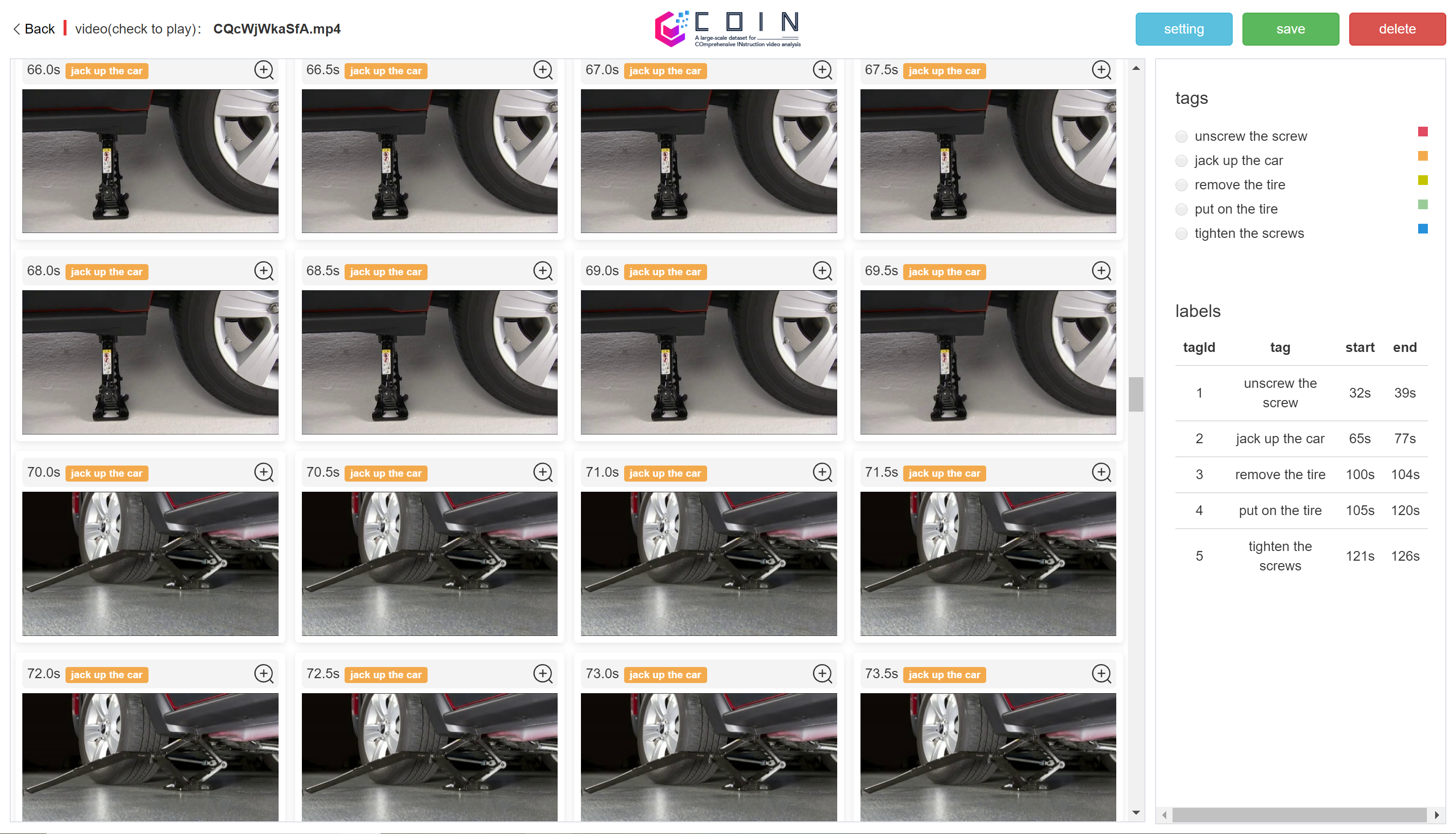}
\caption{The interface of our new developed annotation tool under the \textit{frame mode}.}
\end{figure}

\begin{figure}[tb]
\includegraphics[width = \linewidth]{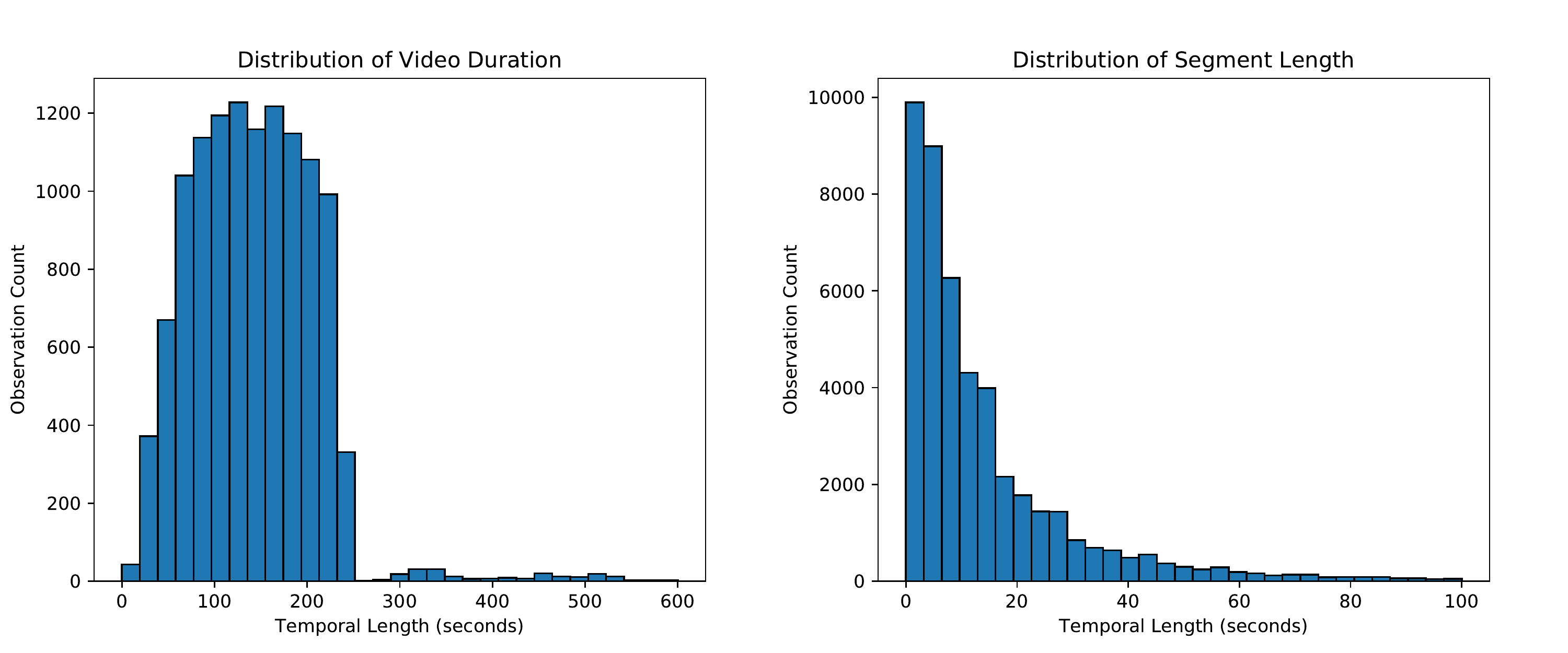}
\caption{
The duration statistics of the videos (left) and segments (right) in the COIN dataset.
}
\end{figure}

\begin{figure*}[tb]
\includegraphics[width = \linewidth]{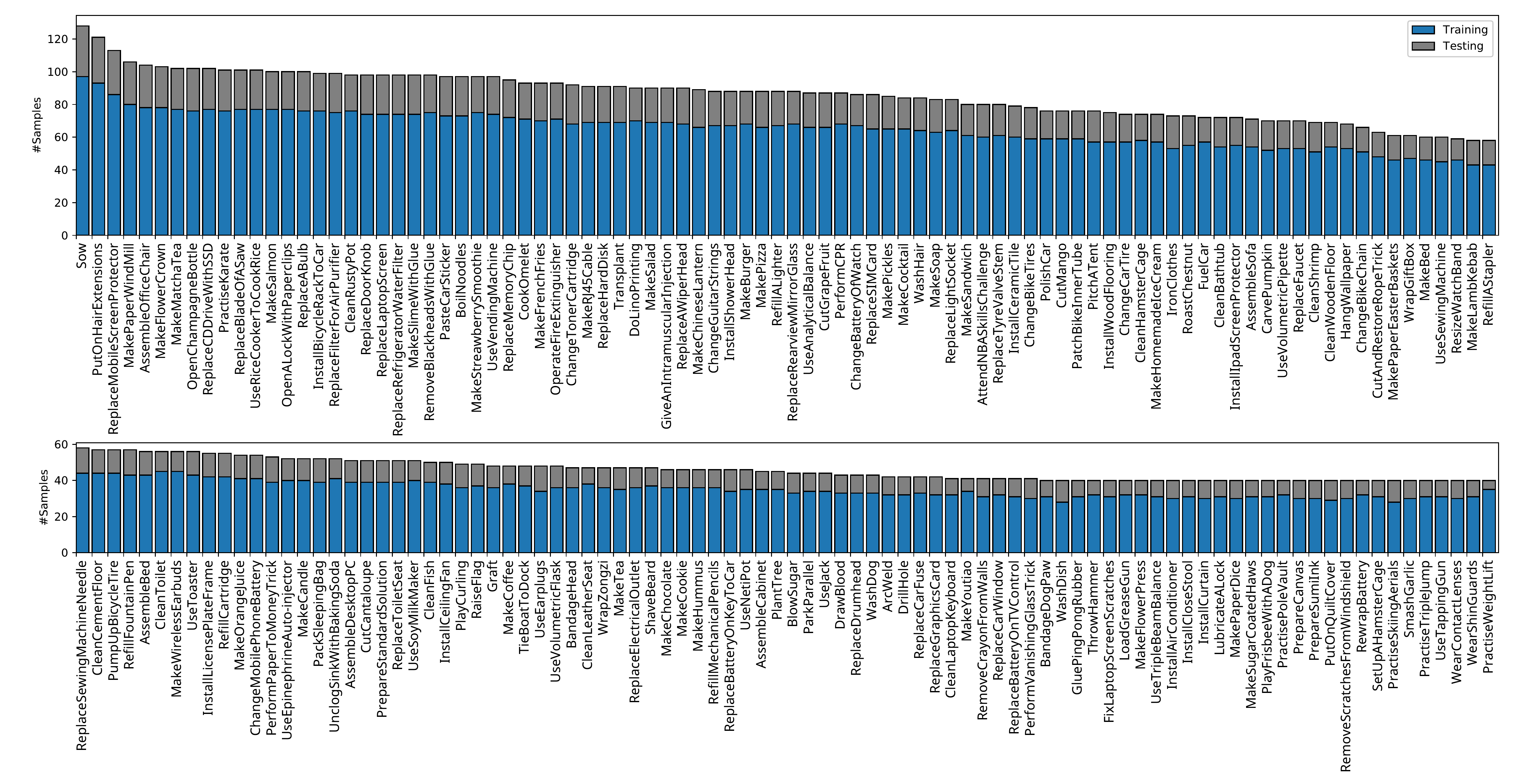}
\caption{
The sample distributions of all the tasks in COIN.
The blue bars and the grey bars indicate the number of training and testing videos in each class respectively. 
}
\end{figure*}

\textbf{Statistics:}
The COIN dataset consists of 11,827 videos related to 180 different tasks, 
which were all collected from YouTube. 
Figure 5 shows the sample distributions among all the task categories.
In order to alleviate the effect of long tails,
we make sure that there are more than 39 videos for each task.
We split the COIN into 9030 and 2797 video samples for training and testing respectively.
Figure 4 displays the duration distribution of videos and segments.
The averaged length of a video is 2.36 minutes.
Each video is labelled with 3.91 step segments, where each segment lasts 14.91 seconds on average. 
In total, the dataset contains videos of 476 hours, with 46,354 annotated segments.

\begin{figure}[tb]
\includegraphics[width = \linewidth]{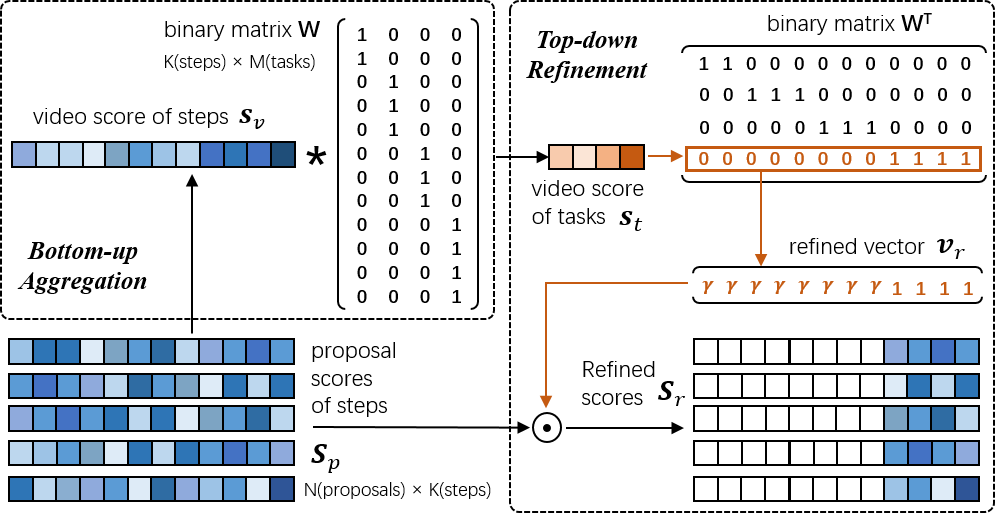}
\caption{Flowchart of our proposed task-consistency method.
During the first \textbf{bottom-up aggregation} stage,
the inputs are a series of scores $\textbf{S}_p=\{\textbf{s}_p^1, ... , \textbf{s}_p^n, ..., \textbf{s}_p^N\}$ of an instructional video,
which denotes the probabilities of each step appearing in the corresponding proposal.
We first aggregate them into a video-based score $\textbf{s}_v$, 
and map it into another score $\textbf{s}_t$ to predict the task label $L$.
At \textbf{top-down refinement} stage,
we generate a refined mask vector $\textbf{v}_r$ based on the task label.
Then we alleviate the weights of other bits in $\textbf{S}_p$ by $\textbf{v}_r$ to ensure the task-consistency.
The refined scores $\textbf{S}_r$ are finally utilized to perform NMS process and output the final results.
}
\end{figure}

\begin{table*}[ht]
\small
\caption{\small Comparisons of the step localization accuracy (\%) on the COIN dataset.} \label{tab:table5}
\setlength{\tabcolsep}{9.5pt}
\centering
\begin{tabular}{l | c c c c c | c c c c c }
\toprule[1.5pt]
 & \multicolumn{5}{|c|}{ mAP @ $\alpha$} &
\multicolumn{5}{|c}{ mAR @ $\alpha$}\\
Method & 0.1 & 0.2 & 0.3 & 0.4 & 0.5
& 0.1 & 0.2 & 0.3 & 0.4 & 0.5\\
\midrule[1.2pt]
Random & 0.03 & 0.03 & 0.02 & 0.01 & 0.01 & 2.57 & 1.79 & 1.36& 0.90 & 0.50\\
\hline
R-C3D\cite{DBLP:conf/iccv/XuDS17} &   9.85 & 7.78 & 5.80 & 4.23 & 2.82  & 36.82 & 31.55 & 26.56 & 21.42 & 17.07 \\
SSN$_{\textit{-RGB}}$\cite{DBLP:conf/iccv/ZhaoXWWTL17} & 19.39 & 15.61 & 12.68 & 9.97 & 7.79 & 50.33 & 43.42 & 37.12 & 31.53 & 26.29\\
SSN$_{\textit{-Flow}}$\cite{DBLP:conf/iccv/ZhaoXWWTL17} &  11.23 & 9.57 & 7.84 & 6.31 & 4.94 & 33.78 & 29.47 & 25.62 & 21.98 & 18.20\\
SSN$_{\textit{-Fusion}}$\cite{DBLP:conf/iccv/ZhaoXWWTL17} &  20.00 & 16.09 & 13.12 & 10.35 & 8.12 & 51.04 & 43.91 & 37.74 & 32.06 & 26.79\\
\hline
R-C3D+TC &  10.32 & 8.25 & 6.20 & 4.54 & 3.08 & 39.25 & 34.22 & 29.09 & 23.71 & 19.24 \\
SSN+TC$_{\textit{-RGB}}$ & 20.15 & 16.79 & 14.24 & 11.74 & 9.33 & 54.05 & 47.31 & 40.99 & 35.11 & 29.17 \\
SSN+TC$_{\textit{-Flow}}$ &  12.11 & 10.29 & 8.63 & 7.03 & 5.52 & 37.24 & 32.52 & 28.50 & 24.46 & 20.58 \\
SSN+TC$_{\textit{-Fusion}}$ &  20.01 & 16.44 & 13.83 & 11.29 & 9.05 & 54.64 & 47.69 & 41.46 & 35.59 & 29.79\\
\bottomrule[1.5pt]
\end{tabular}
\end{table*}

\section{Task-Consistency Analysis}
Given an instructional video, 
one important real-world application is to localize a series of steps to complete the corresponding task.
In this section, we introduce a new proposed task-consistency method for step localization in instructional videos.
Our method is motivated by the intrinsic dependencies of different steps which are associated to a certain task.
For example,
it is unlikely for the steps of ``dig a pit of proper size'' and ``soak the strips into water''
to occur in the same video, because they belong to different tasks of ``plant tree'' and ``make french fries'' respectively.
In another word,
the steps in the same video should be task-consistent to ensure that they belong to the same task.
Figure 6 presents the flowchart of our task-consistency method,
which contains two stages: (1) bottom-up aggregation and (2) top-down refinement.

\textbf{Bottom-up aggregation.}
As our method is built upon the proposal-based action detection methods,
we start with training an existing action detector, \textit{e.g.} SSN~\cite{DBLP:conf/iccv/ZhaoXWWTL17}, on our COIN dataset.
During inference phase, given an input video, we send it into the action detector to produce a series of proposals with their corresponding locations and predicted scores.
These scores indicate the probabilities of each step occuring in the corresponding proposal.
We denote them as $\textbf{S}_p = \{\textbf{s}_p^1, ..., \textbf{s}_p^n, ..., \textbf{s}_p^N\}$, 
where $\textbf{s}_p^n \in R^K$ represents the score of the $n-th$ proposal and $K$ is the number of the total steps.
The goal of the bottom-up aggregation stage is to predict the task labels based on these proposal scores.
To this end, we first aggregate the scores along all the proposals as $\textbf{s}_v = \sum_{n=1}^N \textbf{s}_p^n$,
where $\textbf{s}_v$ indicates the probability of each step appearing in the video.
Then we construct a binary matrix $W$ with the size of $K \times M$
to model the relationship between the $K$ steps and $M$ tasks: 
\begin{eqnarray}
    w_{ij} = 
    \begin{cases}
1, \quad \text{if step $i$ belongs to task $j$}\\
0, \quad \text{otherwise} \\
    \end{cases}
\end{eqnarray}

Having obtained the step-based score $\textbf{s}_v$ and the binary matrix $W$,
we calculate a task-based score as $\textbf{s}_t = \textbf{s}_v * W$. 
This operation is essential to combine the scores of steps belonging to same tasks. 
We choose the index $L$ with the max value in the $\textbf{s}_t$ as the task label of the entire video.

\textbf{Top-down refinement.}
The target of the top-down refinement stage is to refine the original proposal scores with the guidance of the task label.
We first select the $L-th$ row in $W$ as a mask vector $\textbf{v}$,
based on which we define a refined vector as:
\begin{eqnarray}
 \textbf{v}_r = \textbf{v} + \gamma (\textbf{I} - \textbf{v}) .
\end{eqnarray}

Here $\textbf{I}$ is an vector where all the elements equal to 1.
$\gamma$ is an attenuation coefficient to alleviate the weights of the steps which do not belong to the task $L$.
We empirically set  $\gamma$ to be $e^{-2}$ in this paper.
Then, we employ the $\textbf{v}_r$ to mask the original scores $\textbf{s}_p^n$ as follow:
\begin{equation}
    \textbf{s}_r^n = \textbf{s}_p^n \odot \textbf{v}_r ,
\end{equation}
where $\odot$ is the element-wise Hadamard product.
We compute a sequence of scores as $\textbf{S}_r = \{\textbf{s}_r^{1}, ..., \textbf{s}_r^{n}, ..., \textbf{s}_r^{N}\}$.
Based on these refined scores and their locations,
we employ a Non-Maximum Suppression (NMS) strategy to obtain the results of step localization.
In summary, we first predict the task label through the bottom-up scheme,
and refine the proposal scores by the top-down strategy,
hence the task-consistency is guaranteed.

\begin{figure*}[tb]
\includegraphics[width = \linewidth]{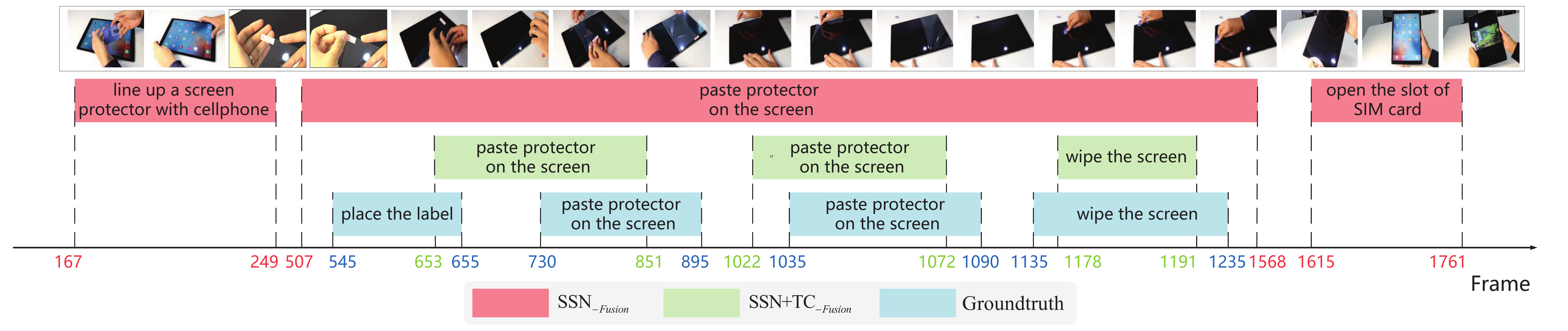}
\caption{Visualization of step localization results. 
The video is associated with the task ``paste screen protector on Pad''.
}
\end{figure*}

\section{Experiments}
In order to provide a benchmark for our COIN dataset, 
we evaluate various approaches under two different settings: step localization and action segmentation.
We also conduct experiments on our task-consistency method under the first setting.
The following describes the details of our experiments and results.

\subsection{Evaluation on Step Localization}
\textbf{Implementation Details.}
In this task, we aim to localize a series of steps and recognize their corresponding labels given an instructional video.
We mainly evaluate the following approaches:
(1) Random. We uniformly segmented the video into three intervals, and randomly assigned the label to each interval.  
(2) R-C3D\cite{DBLP:conf/iccv/XuDS17} and SSN\cite{DBLP:conf/iccv/ZhaoXWWTL17}.
These two methods are state-of-the-arts for action detection, 
which output the same type of results (interval and label for each action instance) with step localization.
For R-C3D, our implementation was built upon the codebase~\cite{rc3dcode}.
Following \cite{DBLP:conf/iccv/XuDS17}, we extracted the RGB frames of each video as the inputs, 
and it took around 3.5 days to train the model on a GTX 1080Ti GPU.
For SSN, 
we used the PyTorch implementation~\cite{ssncode}.
The reported results are based on the inputs of different modalities as:
SSN$_{-RGB}$, SSN$_{-Flow}$ and SSN$_{-Fusion}$. 
Here SSN$_{-Flow}$ adopted the optical flows calculated by ~\cite{zach2007duality},
and SSN$_{-Fusion}$ combined the predicted scores of SSN$_{-RGB}$ and SSN$_{-Flow}$.
(3) R-C3D+TC and SSN+TC. 
In order to demonstrate the advantages of the proposed method to explore the task-consistency in instructional videos,
we further conducted experiments on applying our approach to R-C3D and SSN respectively.

\textbf{Evaluation Metrics:}
As the results of step localization contain time intervals, labels and confidence scores, 
we employed Intersection over Union (IoU) as a basic metric to determine whether a detected interval is positive or not.
The IoU is defined as $|G\cap D| / |G\cup D|$, where G denotes the ground truth action interval and D denotes the detected action interval. 
We followed \cite{DBLP:journals/corr/LiuHLSL17} to calculate \textit{Mean Average Precision (mAP)} and \textit{Mean Average Recall (mAR)}.
The results are reported under the IoU threshold $\alpha$ ranging from 0.1 to 0.5.

\textbf{Results:}
Table 2 presents the compared experimental results,
which reveal great challenges to performing step localization on the COIN dataset.
Even for the state-of-the-art method SSN$_{-Fusion}$, 
it only attains the results of 8.12\% and 26.79\% on mAP@0.5 and mAR@0.5 respectively.
Besides, we observe that R-C3D+TC and SSN+TC consistently improve the performance over the original models,
which illustrates the effectiveness of our proposed method to capture the dependencies among different steps.

We show the visualization results of different methods and ground-truth in Figure 7. 
We analyze an instructional video of the task ``paste screen protector on Pad''.
When applying our task-consistency method, 
we can discard those steps which do not belong to this task, e.g., ``line up a screen protector with cellphone'' and ``open the slot of SIM card'',
hence more accurate step labels can be obtained.
More visualization results are presented in supplementary material.

\subsection{Evaluation on Action Segmentation}
\textbf{Implementation Details:}
The goal of this task is to assign each video frame with a step label. 
We present the results on three types of approaches as follows.
(1) Random. We randomly assigned a step label to each frame. 
(2) Fully-supervised method. We used VGG16 network pretrained on ImageNet,
and finetuned it on the training set of COIN to predict the frame-level label. 
(3) Weakly-supervised approaches. 
In this setting, we evaluated recent proposed Action-Sets\cite{richard2017action}, NN-Viterbi\cite{DBLP:journals/corr/abs-1805-06875} and TCFPN-ISBA\cite{DBLP:journals/corr/abs-1803-10699} without temporal supervision. 
For Action-Sets, only a set of steps within a video is given, while the occurring order of steps are also provided for NN-Viterbi and TCFPN-ISBA. 
We used frames or their representations sampled at 10fps as input.
We followed the default train and inference pipeline of Action-Sets\cite{actionsets_code}, NN-Viterbi\cite{viterbi_code} and TCFPN-ISBA\cite{isba_code}.
However, these methods use frame-wise fisher vector as video representation,
which comes with huge computation and storage cost on the COIN dataset\footnote{ The calculation of fisher vector is based on the improved Dense Trajectory (iDT) representation~\cite{wang2013action}, which requires huge computation cost and storage space.}. 
To address this,
we employed a bidirectional LSTM on the top of a VGG16 network to extract dynamic feature of a video sequence\cite{DBLP:journals/pami/DonahueHRVGSD17}.

\textbf{Evaluation Metrics:}
We adopted frame-wise accuracy (FA), which is a common benchmarking metric for action segmentation.
It is computed by first counting the number of correctly predicted frames, 
and dividing it by the number of total video frames.

\begin{table}[tb]
\small
\caption{\small Comparisons of the action segmentation accuracy (\%) on the COIN dataset.} \label{tab:table5}
\vskip 0.1 in
\centering
\begin{tabular}{l | c | c}
\toprule[1.5pt]
Method & Frame Acc. & Setting\\
\hline
Random & 0.13 & - \\ 
\hline
CNN~\cite{Karen15very} & 25.79& fully-supervised\\
\hline
Action-Sets\cite{richard2017action} &  4.94 & weakly-supervised\\
NN-Viterbi\cite{DBLP:journals/corr/abs-1805-06875} & 21.17 & weakly-supervised\\
TCFPN-ISBA\cite{DBLP:journals/corr/abs-1803-10699} & 34.30 & weakly-supervised\\
\bottomrule[1.5pt]
\end{tabular}
\end{table}

\textbf{Results: }Table 3 shows the experimental results of action segmentation on the COIN dataset. 
Given the weakest supervision of video transcripts without ordering constraint,
Action-Sets\cite{richard2017action} achieves the result of 4.94\% frame accuracy. 
When taking into account the ordering information, 
NN-Viterbi~\cite{DBLP:journals/corr/abs-1805-06875} and TCFPN-ISBA~\cite{DBLP:journals/corr/abs-1803-10699}
outperform Action-Sets with a large margin of 16.23\% and 29.66\% respectively. 
As a fully-supervised method, CNN~\cite{Karen15very} reaches an accuracy 25.79\%, 
which is much higher than Action-Sets.
This is because CNN utilizes the label of each frame to perform classification and the supervision is much stronger than Action-Sets.
However, as the temporal information and ordering constraints are ignored,
the result of CNN is inferior to TCFPN-ISBA.

\subsection{Discussion}
\textbf{What are the hardest and easiest domains for instructional video analysis?}
In order to provide a more in-depth analysis of the COIN dataset, 
we report the performance of SSN+TC$_{-Fusion}$ among the 12 domains of COIN.
Table 4 presents the comparison results,
where the domain ``sports'' achieves the highest mAP of 30.20\%, 
This is because the differences between the ``sports'' steps are more clear,
thus they are easier to be identified.
In contrast, the results of ``gadgets'' and ``science \& craft'' are relatively low.
The reason is that the steps in these two domains usually have higher similarity with each other.
For example, the step ``remove the tape of the old battery'' is similar with the step ``take down the old battery''.
Hence it is harder to localize the steps in these two domains.
We also show the compared performance across different tasks in the supplementary material.

\begin{table}[tb]
\small
\caption{\footnotesize Comparisons of the step localization accuracy (\%) over 12 domains on the COIN dataset. 
We report the results obtained by SSN+TC$_{-Fusion}$ with $\alpha$ = 0.1.}
\vskip 0.1 in
\centering
\begin{tabular}{ c c | c  c  }
\toprule[1.5pt]
Domain & mAP & Domain & mAP \\
\hline
nursing \& caring  &  22.92  & vehicles & 19.07 \\
science \& craft   &  16.59  & electric appliances  & 19.86  \\
leisure \& performance  &  24.32  & gadgets & 17.99  \\
snacks \& drinks  &  19.79  & dishes & 23.76  \\
plants \& fruits  &  22.71  & sports & 30.20  \\
household items  &  19.07  & housework & 20.70  \\
\bottomrule[1.5pt]
\end{tabular}
\end{table}

\textbf{Can the proposed task-consistency method be applied to other instructional video datasets?}
In order to demonstrate the effectiveness of our proposed method,
we further conduct experiments on another dataset called ``Breakfast"\cite{DBLP:conf/cvpr/KuehneAS14},  which is also widely-used for instructional video analysis. 
The Breakfast dataset contains over 1.9k videos with 77 hours of 4 million frames. Each video is labelled with 
 a subset of 48 cooking-related action categories. Following the default setting, we set split 1 as testing set and the other splits as training set.
Similar to COIN,
we employ SSN\cite{DBLP:conf/iccv/ZhaoXWWTL17}, which is a state-of-the-art method for action detection, 
as a baseline method under the setting of step localization.
As shown in Table 5, 
our proposed task-consistency method improves the performance of the baseline model,
which further shows its advantages to model the dependencies of different steps in instructional videos.

\begin{table}[t]
\small
\caption{Comparisons of the step localization accuracy (\%) on the Breakfast dataset. 
The results are all based on the combination scores of RGB frames and optical flows.
} \label{tab:table5}
\vskip 0.1 in
\centering
\begin{tabular}{l | c c c | c c c }
\toprule[1.5pt]
Metrics & \multicolumn{3}{c}{mAP} &
\multicolumn{3}{|c}{mAR}\\
\hline
Threshold & 0.1 &  0.3 & 0.5
& 0.1& 0.3  & 0.5\\
\hline
SSN\cite{DBLP:conf/iccv/ZhaoXWWTL17} &  28.24 & 22.55 & 15.84 & 54.86 & 45.84 & 35.51\\
\hline
SSN+TC  &  28.25 & 22.73 & 16.39 & 55.51 & 47.37 & 36.20 \\
\bottomrule[1.5pt]
\end{tabular}
\end{table}

\begin{table}[tb]
\centering 
\caption{\footnotesize Comparisons of the proposal localization accuracy (\%) with YouCook2 dataset~\cite{DBLP:conf/aaai/ZhouXC18}.
The results are obtained by temporal actionness grouping (TAG) method~\cite{DBLP:conf/iccv/ZhaoXWWTL17} with $\alpha$ = 0.5.}
\small
\begin{tabular}{ c c c|c c c}
\toprule[1.5pt]
& YouCook2 & COIN & & YouCook2 & COIN \\
\hline
mAP & 40.16 & 39.67 & mAR & 54.12 & 56.16\\
\bottomrule[1.5pt]
\end{tabular}
\end{table}

\textbf{Comparison of state-of-the-art performance on existing datasets for video analysis.}
In order to assess the difficulty of COIN, 
we report the performance on different tasks compared with other datasets. 
For \textit{proposal localization}, which is a task defined in~\cite{DBLP:conf/aaai/ZhouXC18} for instructional video analysis,
we evaluated COIN and Youcook2~\cite{DBLP:conf/aaai/ZhouXC18} based on temporal actionness grouping (TAG) approach~\cite{DBLP:conf/iccv/ZhaoXWWTL17}.
From the results in Table 6, we observe that these two datasets are almost equally challenging on this task.
For \textit{video classification} on COIN,
we present the recognition accuracy of 180 tasks, which refer to the second level of the lexicon.
We employed the temporal segment network (TSN) model~\cite{TSN2016ECCV,tsncode},
which is a state-of-the-art method for video classification.
As shown in the Table 7,
the classification accuracy on COIN is 88.02\%,
suggesting its general difficulty in comparison with other datasets.
For \textit{action detection} or \textit{step localization},
we display the compared performances of structured segment networks (SSN) approach~\cite{DBLP:conf/iccv/ZhaoXWWTL17} on COIN and the other three datasets.
The THUMOS14~\cite{THUMOS14} and ActivityNet~\cite{DBLP:conf/cvpr/HeilbronEGN15} are conventional datasets for action detection,
on which the detection accuracies are relatively higher.
%
The Breakfast~\cite{DBLP:conf/cvpr/KuehneAS14} and COIN contain instructional videos with more difficulty.
Hence, the performance on these two datasets are lower.
Especially for our COIN, the results of mAP@0.5 is only 8.12\%. 
We attribute the low performance to two aspects:
(1) The step intervals are usually shorter than action instances, which brings more challenges for temporal localization; 
(2) Some steps in the same tasks are similar,
which carry ambiguous information for the recognition process.
These two phenomena are also common in real-world scenarios,
and future works are encouraged to address these two issues.

\begin{table}[tb]
\footnotesize
\caption{\footnotesize Comparisons of the performance (\%) on different datasets.
The video classification task is evaluated by temporal segment networks (TSN) model~\cite{TSN2016ECCV},
while the action detection task is tested on stuctured segment networks (SSN) method~\cite{DBLP:conf/iccv/ZhaoXWWTL17} with $\alpha$ = 0.5.} \label{tab:table5}
\vskip 0.1 in
\centering
\begin{tabular}{l c | l c}
\toprule[1.5pt]
\multicolumn{2}{c}{\textbf{Video Classification}} & \multicolumn{2}{|c}{\textbf{Action Detection / Step Localization}}  \\
\hline
Dataset & Acc. & Dataset & mAP  \\
\hline
UCF101~\cite{UCF101} & 97.00 & THUMOS14~\cite{THUMOS14} & 29.10\\
ActivityNet v1.3~\cite{DBLP:conf/cvpr/HeilbronEGN15} & 88.30  & ActivityNet v1.3~\cite{DBLP:conf/cvpr/HeilbronEGN15} & 28.30 \\
Kinectics~\cite{DBLP:conf/cvpr/CarreiraZ17} & 73.90 & Breakfast~\cite{DBLP:conf/cvpr/KuehneAS14} & 15.84 \\
\hline
COIN & 88.02 & COIN & 8.12 \\
\bottomrule[1.5pt]
\end{tabular}
\end{table}

\section{Conclusions}
In this paper we have introduced COIN, 
a new large-scale dataset for comprehensive instructional video analysis.
Organized in a rich semantic taxonomy, 
the COIN dataset covers boarder domains and contains more tasks than existing instructional video datasets.
In addition, we have proposed a task-consistency method to explore the relationship among different steps of a specific task.
In order to establish a benchmark,
we have evaluated various approaches under different scenarios on the COIN.
The experimental results have shown the great challenges on the COIN and the effectiveness of our proposed method.

\section*{Acknowledgement}
This work was supported in part by the National Natural Science Foundation of China under Grant U1813218, Grant 61822603, Grant U1713214, Grant 61672306, Grant 61572271 and Meitu Cloud Vision Team.
The authors would like to thank Yongxiang Lian, Jiayun Wang, Yao Li, Jiali Sun and Chang Liu for their generous help.

\section*{Supplementary Material}

\setcounter{section}{0}
\setcounter{table}{0}
\setcounter{figure}{0}

\section{Annotation Time Cost Analysis}
In section 3, we have introduced a toolbox for annotating COIN dataset.
The toolbox has two modes:  frame mode and video mode.
The frame mode is new developed for efficient annotation, while the video mode is frequently used in previous works~\cite{DBLP:conf/iccv/KrishnaHRFN17}.
We have evaluated the annotation time on a small set of COIN, which contains 25 videos of 7 tasks.
Table 1 shows the comparison of annotation time under two different modes.
We observe that the annotation time under the frame mode is only 26.8\% of that under the video mode, 
which shows the advantages of our developed toolbox.

\begin{table}[!h]
\caption{Comparisons of the annotation time cost under two modes. FM indicates the new developed frame mode, and VM represents the conventional video mode.}
\small
\centering
\begin{tabular}{ r  c  r r }
\toprule[1.5pt]
Task & samples  &  FM & VM\\
\hline
Assemble Bed  &  6  & 6:55 & 23:30 \\
Boil Noodles  &  5   & 3:50 & 18:15 \\
Lubricate A Lock   & 2   & 1:23 & 5:29 \\
Make French Fries &  6   & 5:57 &20:24 \\
Change Mobile Phone Battery &  2   & 2:23 & 7:35 \\
Replace A Bulb &  2   & 1:30  & 6:40\\
Plant A Tree &  2   & 1:45  & 6:37\\
\hline
Total & 25 &  23:43 & 88:30  \\
\bottomrule[1.5pt]
\end{tabular}
\end{table}

\begin{figure}[!h]
\includegraphics[width = \linewidth]{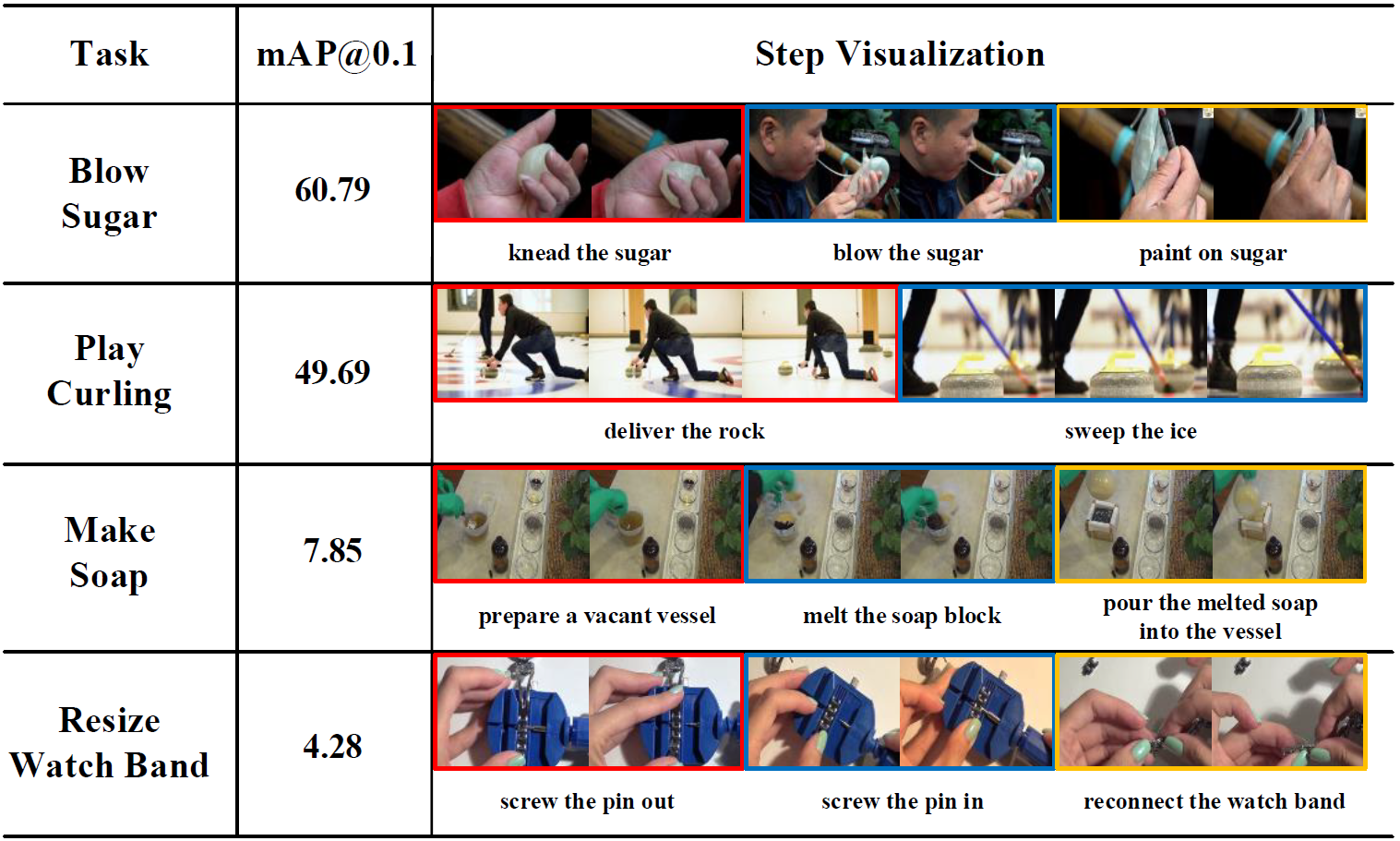}
\caption{
Comparisons of the step localization accuracy (\%) of different tasks. 
We report the results obtained by SSN+TC$_{-Fusion}$ with $\alpha$ = 0.1.
}
\vspace{-3pt}
\end{figure}

\section{Browse Times Analysis}
In order to justify that the selected tasks meet the need of website viewers, 
we display the number of browse times across 180 tasks in Figure 1. 
We searched ``How to'' + name of 180 tasks, e.g., ``How to Assemble Sofa'', on YouTube respectively. 
Then we summed up the browse times of the videos appearing in the first pages (about 20 videos) to get the final results.
``Make French Fries'' is the most-viewed task, which has been browsed $1.7 \times 10^8$ times.
And the browse times per task are $2.3 \times 10^7$ on average.
These results demonstrate the selected tasks of our COIN dataset satisfy the need of website viewers,
and also reveal the practical value of instructional video analysis.

\begin{figure*}[tb]
\includegraphics[width = \linewidth]{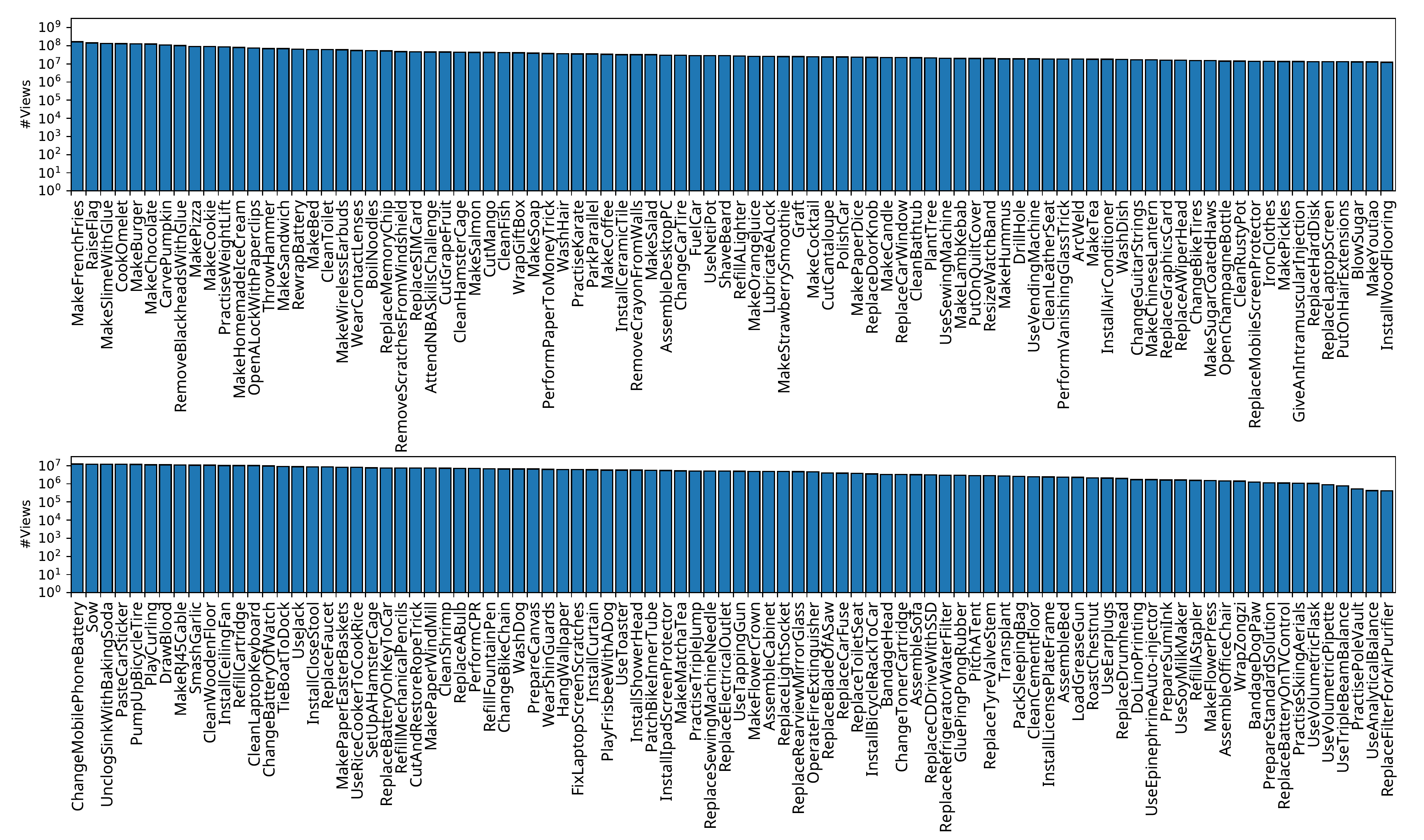}
\caption{The browse time distributions of the selected 180 tasks on YouTube.}
\end{figure*}

\begin{figure*}[tb]
\includegraphics[width = \linewidth]{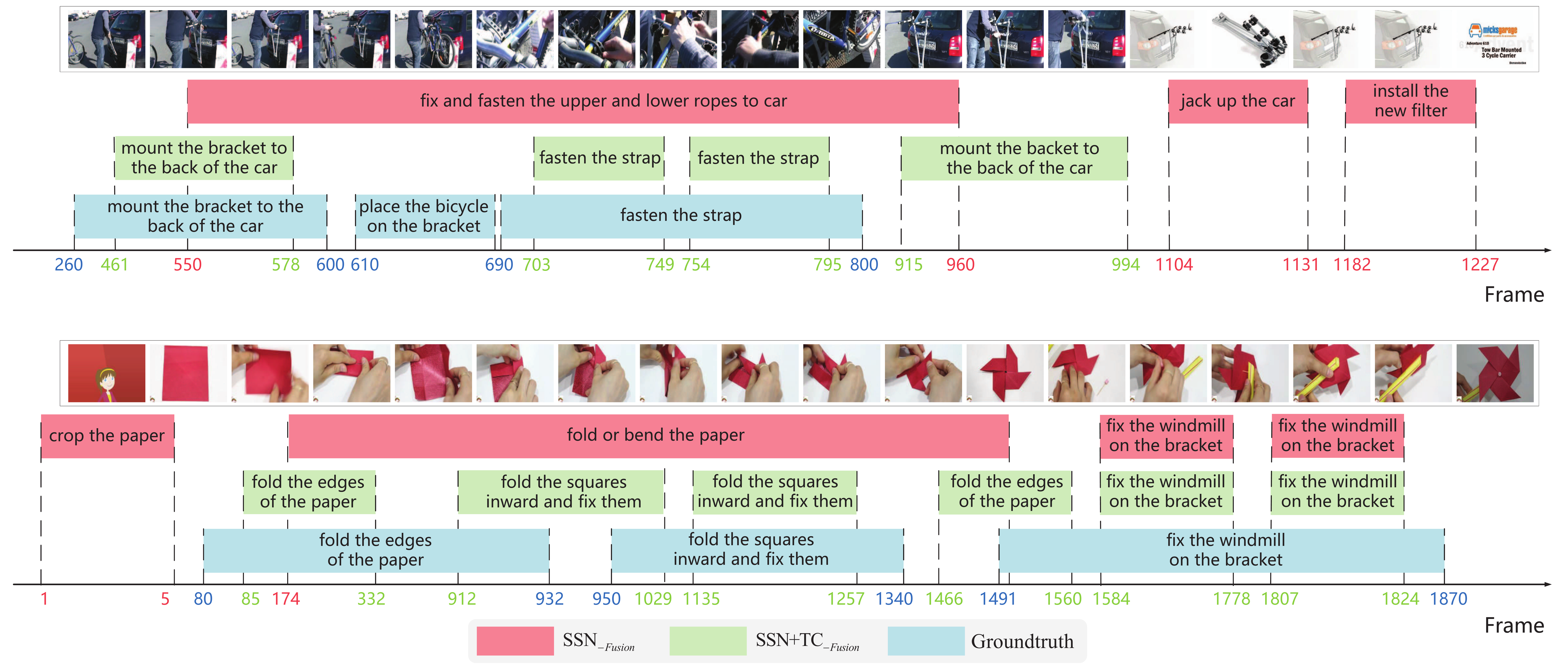}
\caption{Visualization of step localization results. 
The videos are associated with the task ``install the bicycle rack'' and ``make paper windmill''. }
\end{figure*}

\section{Visualization Results}
In section 5.1, we have visualized the step localization results of different methods and ground-truth annotations. 
Figure 2 shows more examples of task ``install bicycle rack'' and ``make paper windmill''. 
When applying our task-consistency method, we can discard the steps which do not belong to the correct task, e.g., ``jack up the car'' in the task ``install the bicycle rack'' and ``crop the paper'' in the task ``make paper windmill''. 
For more visualization results, please see the uploaded video.

\section{Step Localization Results of Different Tasks}
In section 5.3, we have compared the performance across different domains.
Figure 3 shows some examples from 4 different tasks as ``blow sugar'', ``play curling'', ``make soap'' and ``resize watch band''.
They belong to the domain ``sports'',
``leisure \& performance'', ``gadgets'' and ``science and craft'', which are the two of the easiest domains and the two of the hardest domains.
For ``blow sugar'' and ``play curling'',
different steps vary a lot in appearance, thus it is easier to localize them in videos.
For ``make soap'' and ``resize watch band'',
various steps tend to occur in similar scenes, hence the mAP accuracy of these tasks are inferior.

{\small
\bibliographystyle{ieee}
\bibliography{egbib}

\begin{thebibliography}{10}\itemsep=-1pt

\bibitem{actionsets_code}
Codebase of {Action-Sets}.
\newblock \url{ https://github.com/alexanderrichard/action-sets }.

\bibitem{viterbi_code}
Codebase of {NN-Viterbi}.
\newblock \url{ https://github.com/alexanderrichard/NeuralNetwork-Viterbi }.

\bibitem{rc3dcode}
Codebase of {R-C3D}.
\newblock \url{https://github.com/VisionLearningGroup/R-C3D}.

\bibitem{ssncode}
Codebase of {SSN}.
\newblock \url{https://github.com/yjxiong/action-detection}.

\bibitem{isba_code}
Codebase of {TCFPN-ISBA}.
\newblock \url{https://github.com/Zephyr-D/TCFPN-ISBA}.

\bibitem{tsncode}
Codebase of {TSN}.
\newblock \url{https://github.com/yjxiong/temporal-segment-networks}.

\bibitem{howcast}
Howcast.
\newblock \url{https://www.howcast.com}.

\bibitem{howdini}
Howdini.
\newblock \url{https://www.howdini.com}.

\bibitem{wikihow}
Wikihow.
\newblock \url{https://www.wikihow.com}.

\bibitem{DBLP:conf/cvpr/AlayracBASLL16}
J.~Alayrac, P.~Bojanowski, N.~Agrawal, J.~Sivic, I.~Laptev, and
  S.~Lacoste{-}Julien.
\newblock Unsupervised learning from narrated instruction videos.
\newblock In {\em CVPR}, pages 4575--4583, 2016.

\bibitem{DBLP:conf/cvpr/CarreiraZ17}
J.~Carreira and A.~Zisserman.
\newblock Quo vadis, action recognition? {A} new model and the kinetics
  dataset.
\newblock In {\em CVPR}, pages 4724--4733, 2017.

\bibitem{DBLP:journals/corr/LiuHLSL17}
L.~Chunhui, H.~Yueyu, L.~Yanghao, S.~Sijie, and L.~Jiaying.
\newblock {PKU-MMD}: A large scale benchmark for continuous multi-modal human
  action understanding.
\newblock In {\em ACM MM workshop}, 2017.

\bibitem{Damen_2018_ECCV}
D.~Damen, H.~Doughty, G.~Maria~Farinella, S.~Fidler, A.~Furnari, E.~Kazakos,
  D.~Moltisanti, J.~Munro, T.~Perrett, W.~Price, and M.~Wray.
\newblock Scaling egocentric vision: The epic-kitchens dataset.
\newblock In {\em ECCV}, 2018.

\bibitem{DBLP:conf/cvpr/DasXDC13}
P.~Das, C.~Xu, R.~F. Doell, and J.~J. Corso.
\newblock A thousand frames in just a few words: Lingual description of videos
  through latent topics and sparse object stitching.
\newblock In {\em CVPR}, pages 2634--2641, 2013.

\bibitem{DBLP:journals/prl/AvilaLLA11}
S.~E.~F. de~Avila, A.~P.~B. Lopes, A.~da~Luz~Jr., and
  A.~de~Albuquerque~Ara{\'{u}}jo.
\newblock {VSUMM:} {A} mechanism designed to produce static video summaries and
  a novel evaluation method.
\newblock {\em Pattern Recognition Letters}, 32(1):56--68, 2011.

\bibitem{DBLP:conf/cvpr/DengDSLL009}
J.~Deng, W.~Dong, R.~Socher, L.~Li, K.~Li, and F.~Li.
\newblock Imagenet: {A} large-scale hierarchical image database.
\newblock In {\em CVPR}, pages 248--255, 2009.

\bibitem{DBLP:journals/corr/abs-1803-10699}
L.~Ding and C.~Xu.
\newblock Weakly-supervised action segmentation with iterative soft boundary
  assignment.
\newblock In {\em CVPR}, pages 6508--6516, 2018.

\bibitem{DBLP:journals/pami/DonahueHRVGSD17}
J.~Donahue, L.~A. Hendricks, M.~Rohrbach, S.~Venugopalan, S.~Guadarrama,
  K.~Saenko, and T.~Darrell.
\newblock Long-term recurrent convolutional networks for visual recognition and
  description.
\newblock {\em TPAMI}, 39(4):677--691, 2017.

\bibitem{DBLP:journals/pami/ElhamifarSS16}
E.~Elhamifar, G.~Sapiro, and S.~S. Sastry.
\newblock Dissimilarity-based sparse subset selection.
\newblock {\em TPAMI}, 38(11):2182--2197, 2016.

\bibitem{WordNet}
C.~Fellbaum.
\newblock Wordnet: An electronic lexical database.
\newblock {\em Bradford Books}, 1998.

\bibitem{Gu_2018_CVPR}
C.~Gu, C.~Sun, D.~A. Ross, C.~Vondrick, C.~Pantofaru, Y.~Li,
  S.~Vijayanarasimhan, G.~Toderici, S.~Ricco, R.~Sukthankar, C.~Schmid, and
  J.~Malik.
\newblock Ava: A video dataset of spatio-temporally localized atomic visual
  actions.
\newblock In {\em CVPR}, pages 6047--6056, 2018.

\bibitem{DBLP:conf/eccv/GygliGRG14}
M.~Gygli, H.~Grabner, H.~Riemenschneider, and L.~J.~V. Gool.
\newblock Creating summaries from user videos.
\newblock In {\em ECCV}, pages 505--520, 2014.

\bibitem{DBLP:conf/cvpr/HeilbronEGN15}
F.~C. Heilbron, V.~Escorcia, B.~Ghanem, and J.~C. Niebles.
\newblock Activitynet: {A} large-scale video benchmark for human activity
  understanding.
\newblock In {\em CVPR}, pages 961--970, 2015.

\bibitem{DBLP:conf/cvpr/HuangLFN17}
D.~Huang, J.~J. Lim, L.~Fei{-}Fei, and J.~C. Niebles.
\newblock Unsupervised visual-linguistic reference resolution in instructional
  videos.
\newblock In {\em CVPR}, pages 1032--1041, 2017.

\bibitem{Huang_2018_CVPR}
D.-A. Huang, S.~Buch, L.~Dery, A.~Garg, L.~Fei-Fei, and J.~Carlos~Niebles.
\newblock Finding "it": Weakly-supervised reference-aware visual grounding in
  instructional videos.
\newblock In {\em CVPR}, pages 5948--5957, 2018.

\bibitem{THUMOS14}
Y.-G. Jiang, J.~Liu, A.~Roshan~Zamir, G.~Toderici, I.~Laptev, M.~Shah, and
  R.~Sukthankar.
\newblock {THUMOS} challenge: Action recognition with a large number of
  classes.
\newblock \url{http://crcv.ucf.edu/THUMOS14/}, 2014.

\bibitem{DBLP:conf/iccv/KrishnaHRFN17}
R.~Krishna, K.~Hata, F.~Ren, L.~Fei{-}Fei, and J.~C. Niebles.
\newblock Dense-captioning events in videos.
\newblock In {\em ICCV}, pages 706--715, 2017.

\bibitem{DBLP:conf/cvpr/KuehneAS14}
H.~Kuehne, A.~B. Arslan, and T.~Serre.
\newblock The language of actions: Recovering the syntax and semantics of
  goal-directed human activities.
\newblock In {\em CVPR}, pages 780--787, 2014.

\bibitem{ATUS}
U.~D. of~Labor.
\newblock American time use survey.
\newblock 2013.

\bibitem{DBLP:journals/tip/PandaMR17}
R.~Panda, N.~C. Mithun, and A.~K. Roy{-}Chowdhury.
\newblock Diversity-aware multi-video summarization.
\newblock {\em TIP}, 26(10):4712--4724, 2017.

\bibitem{faster-rcnn}
S.~Ren, K.~He, R.~B. Girshick, and J.~Sun.
\newblock Faster {R-CNN:} towards real-time object detection with region
  proposal networks.
\newblock In {\em NIPS}, pages 91--99, 2015.

\bibitem{richard2017action}
A.~Richard, H.~Kuehne, and J.~Gall.
\newblock Action sets: Weakly supervised action segmentation without ordering
  constraints.
\newblock In {\em CVPR}, pages 5987--5996, 2018.

\bibitem{DBLP:journals/corr/abs-1805-06875}
A.~Richard, H.~Kuehne, A.~Iqbal, and J.~Gall.
\newblock Neuralnetwork-viterbi: {A} framework for weakly supervised video
  learning.
\newblock In {\em CVPR}, pages 7386--7395, 2018.

\bibitem{Nadolski2005Optimizing}
N.~RJ, K.~PA, and van Merri\"enboer~JJ.
\newblock Optimizing the number of steps in learning tasks for complex skills.
\newblock {\em British Journal of Educational Psychology}, 75(2):223--\ 237,
  2005.

\bibitem{DBLP:conf/cvpr/RohrbachAAS12}
M.~Rohrbach, S.~Amin, M.~Andriluka, and B.~Schiele.
\newblock A database for fine grained activity detection of cooking activities.
\newblock In {\em CVPR}, pages 1194--1201, 2012.

\bibitem{Sener_2015_ICCV}
O.~Sener, A.~R. Zamir, S.~Savarese, and A.~Saxena.
\newblock Unsupervised semantic parsing of video collections.
\newblock In {\em ICCV}, pages 4480--4488, 2015.

\bibitem{Karen15very}
K.~Simonyan and A.~Zisserman.
\newblock Very deep convolutional networks for large-scale image recognition.
\newblock In {\em ICLR}, pages 1--14, 2015.

\bibitem{DBLP:conf/cvpr/SongVSJ15}
Y.~Song, J.~Vallmitjana, A.~Stent, and A.~Jaimes.
\newblock Tvsum: Summarizing web videos using titles.
\newblock In {\em CVPR}, pages 5179--5187, 2015.

\bibitem{UCF101}
K.~Soomro, A.~Zamir, and M.~Shah.
\newblock Ucf101: A dataset of 101 human actions classes from videos in the
  wild.
\newblock {\em Technical Report CRCV-TR-12-01, University of Central Florida},
  2012.

\bibitem{DBLP:conf/huc/SteinM13}
S.~Stein and S.~J. McKenna.
\newblock Combining embedded accelerometers with computer vision for
  recognizing food preparation activities.
\newblock In {\em UbiComp}, pages 729--738, 2013.

\bibitem{DBLP:conf/dicta/ToyerCHG17}
S.~Toyer, A.~Cherian, T.~Han, and S.~Gould.
\newblock Human pose forecasting via deep markov models.
\newblock In {\em {DICTA} ,2017}, pages 1--8, 2017.

\bibitem{DBLP:conf/iccv/TranBFTP15}
D.~Tran, L.~D. Bourdev, R.~Fergus, L.~Torresani, and M.~Paluri.
\newblock Learning spatiotemporal features with 3d convolutional networks.
\newblock In {\em ICCV}, pages 4489--4497, 2015.

\bibitem{wang2013action}
H.~Wang and C.~Schmid.
\newblock Action recognition with improved trajectories.
\newblock In {\em ICCV}, pages 3551--3558, 2013.

\bibitem{TSN2016ECCV}
L.~Wang, Y.~Xiong, Z.~Wang, Y.~Qiao, D.~Lin, X.~Tang, and L.~{Val Gool}.
\newblock Temporal segment networks: Towards good practices for deep action
  recognition.
\newblock In {\em ECCV}, 2016.

\bibitem{DBLP:conf/iccv/XuDS17}
H.~Xu, A.~Das, and K.~Saenko.
\newblock {R-C3D:} region convolutional 3d network for temporal activity
  detection.
\newblock In {\em ICCV}, pages 5794--5803, 2017.

\bibitem{DBLP:conf/cvpr/XuMYR16}
J.~Xu, T.~Mei, T.~Yao, and Y.~Rui.
\newblock {MSR-VTT:} {A} large video description dataset for bridging video and
  language.
\newblock In {\em CVPR}, pages 5288--5296, 2016.

\bibitem{Yu_2018_CVPR}
H.~Yu, S.~Cheng, B.~Ni, M.~Wang, J.~Zhang, and X.~Yang.
\newblock Fine-grained video captioning for sports narrative.
\newblock In {\em CVPR}, pages 6066--6015, June 2018.

\bibitem{zach2007duality}
C.~Zach, T.~Pock, and H.~Bischof.
\newblock A duality based approach for realtime tv-l1 optical flow.
\newblock In {\em DAGM}, pages 214--223, 2007.

\bibitem{DBLP:conf/eccv/ZhangCSG16}
K.~Zhang, W.~Chao, F.~Sha, and K.~Grauman.
\newblock Video summarization with long short-term memory.
\newblock In {\em ECCV}, pages 766--782, 2016.

\bibitem{DBLP:conf/iccv/ZhaoXWWTL17}
Y.~Zhao, Y.~Xiong, L.~Wang, Z.~Wu, X.~Tang, and D.~Lin.
\newblock Temporal action detection with structured segment networks.
\newblock In {\em ICCV}, pages 2933--2942, 2017.

\bibitem{DBLP:conf/bmvc/ZhouLC18}
L.~Zhou, N.~Louis, and J.~J. Corso.
\newblock Weakly-supervised video object grounding from text by loss weighting
  and object interaction.
\newblock In {\em BMVC}, page~50, 2018.

\bibitem{DBLP:conf/aaai/ZhouXC18}
L.~Zhou, C.~Xu, and J.~J. Corso.
\newblock Towards automatic learning of procedures from web instructional
  videos.
\newblock In {\em AAAI}, pages 7590--7598, 2018.

\bibitem{DBLP:journals/corr/abs-1804-00819}
L.~Zhou, Y.~Zhou, J.~J. Corso, R.~Socher, and C.~Xiong.
\newblock End-to-end dense video captioning with masked transformer.
\newblock In {\em CVPR}, pages 8739--8748, 2018.

\end{thebibliography}
}

\end{document}